\documentclass{article}

\usepackage{PRIMEarxiv}

\usepackage[utf8]{inputenc} 
\usepackage[T1]{fontenc}    
\usepackage{hyperref}       
\usepackage{url}            
\usepackage{booktabs}       
\usepackage{amsfonts}       
\usepackage{nicefrac}       
\usepackage{microtype}      
\usepackage{lipsum}
\usepackage{fancyhdr}       
\usepackage{graphicx}       
\graphicspath{{media/}}     
\usepackage{cleveref}
\usepackage{siunitx}
\usepackage{wrapfig}

\title{Neural spell-checker: Beyond words with synthetic data generation
\thanks{\textit{\underline{Citation}}: 
Matej Klemen, Martin Božič, Špela Arhar Holdt, and Marko Robnik-Šikonja. 2024. Neural Spell-Checker: Beyond Words with Synthetic Data Generation. In Text, Speech, and Dialogue: 27th International Conference, TSD 2024, Proceedings, Part I., 85–96. \url{https://doi.org/10.1007/978-3-031-70563-2_7}} 
}

\author{
  Matej Klemen\textsuperscript{1+}, Martin Božič\textsuperscript{1+}, Špela Arhar Holdt\textsuperscript{2}, Marko Robnik-Šikonja\textsuperscript{1} \\
  \textsuperscript{1} University of Ljubljana, Faculty of Computer and Information Science \\
  Večna pot 113, 1000 Ljubljana, Slovenia \\
  \texttt{\{matej.klemen, martin.bozic, marko.robnik\}@fri.uni-lj.si} \\
  \textsuperscript{2} University of Ljubljana, Faculty of Arts, \\   Aškerčeva cesta 2, 1000 Ljubljana, Slovenia \\
  \texttt{spela.arharholdt@ff.uni-lj.si}
   \\
}

\begin{document}
\maketitle
\def\thefootnote{+}\footnotetext{Equal contribution.}
\def\thefootnote{\arabic{footnote}}

\begin{abstract}
Spell-checkers are valuable tools that enhance communication by identifying misspelled words in written texts. Recent improvements in deep learning, and in particular in large language models, have opened new opportunities to improve traditional spell-checkers with new functionalities that not only assess spelling correctness but also the suitability of a word for a given context. 
In our work, we present and compare two new spell-checkers and evaluate them on synthetic, learner, and more general-domain Slovene datasets. The first spell-checker is a traditional, fast, word-based approach, based on a morphological lexicon with a significantly larger word list compared to existing spell-checkers. The second approach uses a language model trained on a large corpus with synthetically inserted errors. We present the training data construction strategies, which turn out to be a crucial component of neural spell-checkers. Further, the proposed neural model significantly outperforms all existing spell-checkers for Slovene in both precision and recall.
\end{abstract}


\section{Introduction}
Correct spelling enhances the clarity, effectiveness, comprehensibility, and consistency of written communication.
Spell-checkers are tools that detect incorrectly spelled words, and, optionally, suggest their correct versions. Spelling error detection and correction are mostly considered separate problems, although they are sometimes tackled jointly. In this work, we focus on spelling error detection without suggesting corrections.

Existing work on spelling error detection can be divided into traditional non-neural and recent neural approaches. The former approaches validate the correctness of words by comparing them against a reference set of correctly spelled words or their statistics. The neural approaches learn from large amounts of texts corrupted in an opaque way. 
The neural approaches typically perform better as they can learn complex text interactions and detect errors beyond misspelled words, e.g., words that should be split into two or merged into one. However, they are computationally more demanding due to the use of large models and may be harder to use in practical applications such as text editing software.
In this work, we introduce two new spell-checkers for the Slovene language: a non-neural morphological lexicon-based approach (SloSpell) and a neural approach (SloNSpell).  These tools support both computationally less restrictive and more restrictive settings; however, even SloNSpell is designed to be computationally efficient and usable for most contexts.

Most existing work on spelling error detection has focused on broadly-spoken languages such as Chinese and English. However, some coverage exists for a relatively large pool of languages, including less-resourced ones.
In our work, we focus on the morphologically rich Slovene language, for which four off-the-shelf spell-checkers exist: the commercial, rule-based tool Amebis Besana, the freemium open source  LanguageTool, the open source Hunspell used by many open source applications such as LibreOffice, and the Hunspell-based Loris.
Besana is a rule-based grammar checker that can detect and correct various spelling and grammatical issues, including misplaced punctuation. Hunspell is a dictionary-based tool for spell-checking with support for Slovene. The word list used in this tool was released in 2006 and was compiled from a significantly smaller data source than is available today. 
Traditional spell-checking systems take into account limited surrounding context and are unlikely to identify instances of spelling errors that are formally the same as another valid word form (e.g., 'came' instead of 'cane').
In our work, we first present the traditional dictionary-based system SloSpell with an updated dictionary of valid word forms based on a morphological lexicon. 
Second, we present the SloNSpell neural system capable of handling more complex spelling errors.

We make the following contributions:
\begin{itemize}
     \item \emph{SloNSpell}, an efficient neural spell-checker based on the BERT transformer model, together with methodology and data generators for the synthetic construction of spelling errors.
   \item \emph{SloSpell}, classical spell-checker, using by far the largest word list for Slovene, based on the Sloleks 3.0 morphological lexicon \cite{sloleks3}. 
    \item \emph{Evaluation} on synthetic and authentic language data showing strengths and weaknesses of the proposed approaches together with their comparison with existing baselines. The evaluation includes two scenarios: texts written by young learners, sampled in the \v{S}olar-Eval dataset \cite{solar-eval}, and texts written by professional adult writers, sampled in the Lektor-Spelling dataset.
\end{itemize}
We publish the source code and fine-tuned model used in our experiments under permissible open source licenses and release the neural spell-checker in an online interface\footnote{The code is available at \href{https://github.com/matejklemen/slonspell}{https://github.com/matejklemen/slonspell} and the model at \href{https://hf.co/cjvt/SloBERTa-slo-word-spelling-annotator}{https://hf.co/cjvt/SloBERTa-slo-word-spelling-annotator}}.

The remainder of this paper is structured as follows. In \Cref{sec:related-work}, we outline the development of spell-checkers and existing work in Slovene. In \Cref{sec:slospell}, we describe the morphological lexicon Sloleks and how the classical spell-checker is built on top of it.
In \Cref{sec:slonspell}, we describe the neural approach based on the BERT model and synthetically generated errors.
In \Cref{sec:evaluation}, we describe the evaluation protocol and compare our systems against strong baselines. Last, in Section \ref{sec:conclusion}, we provide conclusions and possibilities for future research.

\section{Related work}
\label{sec:related-work}

We review related work that focuses on either isolated spelling error detection or spelling error correction, which encompasses the detection step. Existing work can be divided into classical and neural approaches. 
Non-neural approaches dominated earlier research but are nowadays rare. Nevertheless, these approaches remain in use due to their computational efficiency and practical usability. Below, we only review recent, neural approaches. 

Neural approaches, by definition, use neural networks, in particular language models (LMs), to test the validity of a given word. For example, Li et al. \cite{dcspell-chinese-electra} use the ELECTRA \cite{clark2020electra} discriminator model to detect incorrect Chinese characters. As spelling errors are commonly characterized by external factors such as pronunciation patterns, some approaches inject additional knowledge into LMs to improve detection accuracy. For example, Ji et al. \cite{ji2021-spellbert} include radical and Pinyin information as additional visual and phonetic features in Chinese spelling error detection.

Although most existing research focuses on broadly spoken languages such as Chinese and English, some coverage exists for a relatively large pool of languages, including less-resourced ones such as Croatian \cite{croatian-err-corr} and Urdu \cite{urdu-spelling-err-det-corr}. For Croatian, the authors train an XLM-RoBERTa model on a Croatian dataset with synthetically generated errors.

\section{SloSpell: a spell-checker using a morphological lexicon}
\label{sec:slospell}

To construct a pattern-matching spell-checker, we need a high-quality, comprehensive list of valid word forms. For morphologically rich languages, the number of word forms multiple times exceeds the number of word lemmas -- for Slovene, approximately nine times. 
We use the open-source morphological lexicon Sloleks 3.0 \cite{sloleks3}, containing all word forms for a large set of lexemes, part of which are manually validated.

In this section, we first describe the morphological lexicon and then describe how we built the spell-checker on top of it. 


\textbf{Sloleks 3.0} is a Slovene morphological lexicon containing \num{365340} entries, identifiable by their lemmas. Version 3.0 expands version 2.0 with approximately \num{265000} new lemmas that frequently occur in reference corpora. In addition to the lemma, each entry contains its part of speech category, inflected word forms, and morphosyntactic information annotated automatically following the MULTEXT-East specifications for Slovene \cite{erjavec2003-mte}. 
The entire Sloleks 2.0 was manually validated. Among the new lemmas in Sloleks 3.0, the verbs, adjectives, adverbs, and common nouns were manually validated, while the accentuated forms were automatically generated. The total number of unique word forms in Sloleks 3.0 is \num{3028666}. 

In our \textbf{lexicon-based spell-checker}, the input text is first tokenized into words, and each word is processed in isolation without context.
For each word, several checks are performed to circumvent some limitations due to the nature of the lexicon. The word is treated as correct if it is a number or a numeral, a URL, a punctuation, or an otherwise special symbol.
These exceptions are not covered by the lexicon either because they are not ``normal'' words or because they can take on infinitely many forms.
If none of the stated conditions are true, the word is looked up in the Sloleks lexicon, implemented as a hash map: if it exists, it is considered correct; otherwise, it is flagged as incorrect.

One advantage of using this simple architecture is that future versions of the morphological lexicon can replace the existing one without further adaptations.

\section{SloNSpell: a neural spell-checking approach}
\label{sec:slonspell}


We design an approach using the Slovene BERT model SloBERTa. We first describe the specifics of fine-tuning the SloBERTa model in Section \ref{sec:misspelling_finetuning}, followed by synthetic datasets for different types of spelling errors in Section \ref{sec:misspelling_data}.

\subsection{The BERT model fine-tuning}
\label{sec:misspelling_finetuning}

In this section, we describe the fine-tuning of the Slovene SloBERTa \cite{sloberta} model for the identification of misspelled words in a text. As BERT models use only the encoder part of the transformer architecture \cite{Vaswani2017}, their inference is parallel and considerably faster than generative models, such as T5 or GPT, that generate one output at a time.
The dataset we created for training the SloBERTa model is described next, in Section \ref{sec:misspelling_data}.

To fine-tune the SloBERTa model, we introduce a special mask token after each word in a model input vector. This ensures that the prediction always corresponds to a word, even if it is split into multiple tokens by the tokenizer. At the same time, we construct a vector with labels indicating if the word before the mask token is spelled correctly (0), incorrectly (1), needs to be combined with another word (2), or should be split into two words (3). An example of the input vector paired with its corresponding labels is provided in Table \ref{masked_sentence_table_identifying}.

\begin{table}[htb]
\centering
\caption{A training example of a masked sentence (split into two rows) with labels used for the SloBERTa detector of misspelled words. The word \textquotedblleft Vosil\textquotedblright\ (\textquotedblleft Drving\textquotedblright) is misspelled, indicated by "1" in the labels vector. Words \textquotedblleft av\textquotedblright\ (\textquotedblleft ca\textquotedblright) and \textquotedblleft to\textquotedblright\ (\textquotedblleft r\textquotedblright) should be combined, indicated by "2" in the labels vector. Using the BERT model, we focus on the output tokens that are adjacent to the \textless mask\textgreater\ tokens, ignoring all other output tokens, when calculating the loss. The number of labels corresponds to the number of \textless mask\textgreater\ tokens. Labels are tokens with indices 0, 1, and 2.}
\label{masked_sentence_table_identifying}
\begin{tabular}{ l|l|l } 
 masked sentence & labels & translated masked sentence \\ 
 \hline
 Mečka \textless mask\textgreater\ spi \textless mask\textgreater\ na & 1 0 & \textit{Cot \textless mask\textgreater\ sleeps \textless mask\textgreater\ on } \\
 \textless mask\textgreater\ tip \textless mask\textgreater\ kovnici \textless mask\textgreater\ & 0 2 2 & \textless mask\textgreater\ \textit{key \textless mask\textgreater\ board \textless mask\textgreater\ } \\
 \hline
\end{tabular}
\end{table}

\newpage
\subsection{Synthetic data for misspelling detection}
\label{sec:misspelling_data}

\begin{figure}
    \centering
    \includegraphics[width=0.50\textwidth]{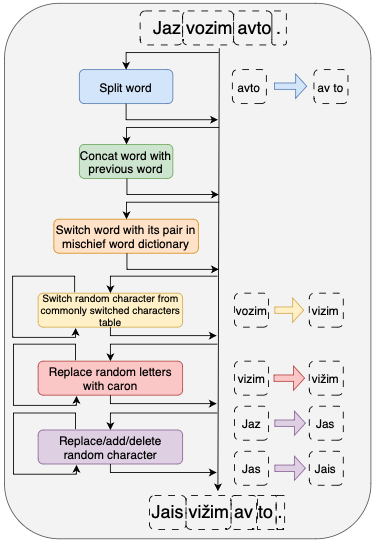}
    \caption{The process of synthetic dataset generation used to train the SloNSpell model. The last three modification methods can be applied multiple times.}
    \label{fig:sloNspell-datapreparation}
\end{figure}

To prepare the synthetic training dataset, we initially segment sentences from the Gigafida 2.0 corpus \cite{krek-etal-2020-gigafida} and group them to form sentence groups, ensuring that their total length does not exceed $128$ tokens. In each sentence group, we identified individual words and randomly modified some of them. We introduced six different word modification methods. The first component splits a given word into two words, the second concatenates two words, while the input and output of the remaining components are individual words. We show the outline of synthetic data preparation in Figure \ref{fig:sloNspell-datapreparation}.

Each word modification method is selected with a probability determined by preliminary experiments on separate synthetically generated data; we tested multiple probabilities in the range from 1\% to 10\%. In the process of synthetic data generation, we aim to generate realistic errors and preserve sentence structure and meaning. Therefore, if a certain modification method is selected, we apply it and then allow for its further selections with the same probability. This simulates multiple spelling errors in the same word. \\

\vspace{-5mm}
\subsubsection{Word split generator}
\label{sec:word-separation}
The generator selects a random position within a word as a candidate for inserting a space. In 99\% of cases, we insert a space only if both parts of the split are recognized as valid words in the Sloleks lexicon (see \Cref{sec:slospell}). In 1\% of cases, we introduce the space without such a validation. 

Within the synthetic dataset for error detection, we apply the word split process with probability $p_{word\_split}$ set to 3\% by default.

\subsubsection{Generating concatenated words}
\label{sec:word-concatenation}
Conceptually similar to splitting words, we concatenate them with the same default probability $p_{conc}=3\%$. We randomly select two words. With probability $p_{conc\_exists}=99\%$, we combine these words if the concatenated word exists in the Sloleks lexicon, and with the probability $1 - p_{conc\_exists}=1\% $ we concatenate them unconditionally. 

%

\subsubsection{Introducing commonly misspelled words}
\label{sec:mischief-word}
We collected words that commonly deviate from their conventional spelling (also called \textquotedblleft mischief words\textquotedblright ), in the Mischief list. We gathered these words from textbooks used in Slovene primary and secondary schools. 

For each word, we check if it exists in this list and replace it with its misspelled version with probability $p_{mischief}$ set to 10\% by default.

\vspace{-3mm}
\subsubsection{Generating commonly switched characters}
\label{sec:mistyped-characters}



We identified commonly switched or replaced characters in mischief words and introduced common character replacements. The switches mostly originate from the similarity of phonemes. 
We select a word with probability $p_{switch\_chr}$ set to 70\% by default. In the selected word, we select four random characters, and if they match one of the characters in the table, we switch them.

Common character switches in Slovene, applied bidirectionally, include: $n \leftrightarrow nj$, $l \leftrightarrow lj$, $t \leftrightarrow d$, $v \leftrightarrow u$, $u \leftrightarrow el$, $i \leftrightarrow j$, $k \leftrightarrow kj$, $k \leftrightarrow h$, $k \leftrightarrow g$, $s \leftrightarrow z$, $p \leftrightarrow b$, $\check{s} \leftrightarrow \check{z}$, $v \leftrightarrow l$, $u \leftrightarrow l$, $t \leftrightarrow tj$, and $i \leftrightarrow ij$.

\subsubsection{Replacement of letters with a caron}
\label{sec:silencers}
The Slovene alphabet contains three letters with a caron: \textit{č}, \textit{š}, and \textit{ž}.
In handwriting, the caret is sometimes forgotten, and in typing, these characters are missing in the English keyboards, therefore they are sometimes substituted with their counterparts without the caron, i.e. c, s, and z.

To generate this type of synthetic error, we select each word containing these characters with probability $p_{caron}$ set by default to $5\%$ and rewrite it by switching them with their non-caron counterparts.

\subsubsection{Replacement of random characters}
\label{sec:characters}

In this error generator, we modify words by switching, replacing, and inserting randomly chosen characters. We sequentially apply each of the five possible modifications to every word in the text: With probability $p_{vowel}= 5\%$, we select a random vowel and swap it with another random vowel. With probability $p_{consonant}= 5\%$, we switch a randomly selected consonant with another random consonant. With probability $p_{subst\_chr}= 2\%$, we substitute a random character with another character from the Slovene alphabet. With probability $p_{del\_chr}= 4\%$, we remove a randomly selected character from a word. With probability $p_{insert\_chr}= 3\%$, we insert a random character at a random position within a word.

\vspace{-3mm}
\subsubsection{Training dataset}
\label{sec:slonspell_training_dataset}

The final training dataset for the SloNSpell model includes \num{31682971} words distributed across \num{297041} sentences. When tokenized with the SloBERTa model tokenizer, the entire training corpus contains \num{39143264} tokens.

\section{Experimental evaluation}
\label{sec:evaluation}

In this section, we describe the evaluation of the proposed spell-checkers. We first describe the evaluation datasets and baseline systems, then move on to the quantitative and qualitative performance analysis.

\subsection{Used evaluation datasets}

We utilize three datasets that sample various types of texts across different evaluation settings: synthetic data and authentic texts from both young learners and adult professional writers. Their summary is displayed in Table \ref{tab:dataset-summary}.

\begin{table}[htb]
\caption{Summary of the evaluation datasets for spelling detection. }
\label{tab:dataset-summary}
\centering
\begin{tabular}{lcrrc}
\toprule
Dataset & text source & \#words & \#sentences & \% errors \\
\midrule
Synthetic-Eval & generated & \num{53990} & \num{959} & $5.29$ \\ 
Šolar-Eval & learners & \num{60271} & \num{740} & $1.27$ \\
Lektor-Spelling & professionals & \num{57336} & \num{1080} & $1.76$ \\
\bottomrule
\end{tabular}
\end{table}

\subsubsection{Synthetic-Eval.} We construct the Synthetic-Eval dataset following the same methodology used for the SloNSpell training dataset, as outlined in Section \ref{sec:misspelling_data}. This involves randomly selecting sentences from the Gigafida 2.0 corpus \cite{krek-etal-2020-gigafida}, ensuring they are not already part of the model's training data. However, in contrast to the training corpus, where we use larger error probabilities to promote learning, in the evaluation, we reduce the likelihood of specific errors to one-eighth of their original values.  This adjustment results in a lower frequency of mistakes so that the error proportion in the Synthetic-Eval dataset is halfway between the error proportion in the two authentic language datasets and the error proportion used during training.
The Synthetic-Eval dataset comprises a total of \num{57336} words, which are distributed across \num{959} sentences.

\subsubsection{Šolar-Eval}
Šolar-Eval\cite{solar-eval} is a dataset containing $109$ essays written by Slovene primary and secondary school students. The essays are sampled from the Šolar 3.0\cite{solar3} corpus which contains essays corrected by school teachers.
As the focus of authentic teacher corrections is on providing educational feedback and not necessarily on annotation completeness, Šolar-Eval was manually annotated by linguist researchers \cite{gantar2023solar-eval} to ensure completeness and consistency and enable its use in the evaluation of grammatical error correction tools. 
The resource is annotated with a diverse range of error categories, including spelling, morphology, vocabulary, syntax, and orthography; in our evaluation, we only consider spelling errors. 

\subsubsection{Lektor-Spelling}

Due to a lack of reference evaluation datasets for Slovene spell-checking, we decided to create a new one for our task. We named this dataset Lektor-Spelling and included 1080 examples, randomly sampled from the Lektor corpus\cite{lektor}, known for containing nonliterary texts together with corrections made by proofreaders. The annotation of Lektor-Spelling uses a similar but not fully interchangeable error category system as that used in Šolar. Following the protocol of Šolar-Eval, our evaluation dataset includes annotated spelling errors, as well as errors related to the conjoining or separating of words and word capitalization. The dataset was cleaned of irrelevant segments (e.g., non-Slovene texts or bibliographical references) and double-checked for consistency of error annotation by two independent annotators.

\subsection{Baseline systems}
As currently there exists no neural spell-checker for Slovene, we utilize three baseline systems: two classical baselines (HunSpell and LanguageTool), as well as ChatGPT 4 as a proxy for neural spell-checker. We present their short description below.

\subsubsection{HunSpell\textsubscript{SL}}
HunSpell is an open-source spell-checker with support for multiple languages used by applications such as LibreOffice and Mozilla Firefox.
Support for Slovene was added in 2006\footnote{Prepared by Amebis, Tomaž Erjavec, Aleš Košir, and Primož Peterlin: \href{https://cgit.freedesktop.org/libreoffice/dictionaries/tree/sl\_SI}{https://cgit.freedesktop.org/libreoffice/dictionaries/tree/sl\_SI}} -- its dictionary contains \num{246856} entries.

\subsubsection{LanguageTool\textsubscript{SL}}
LanguageTool is an open-source style and grammar correction tool supporting multiple languages, including Slovene. It uses the dictionary from HunSpell and additional rules covering certain error patterns such as the use of duplicate words, unclosed parentheses, etc. It has a free and a premium version -- in our work, we use the free version via the API.


\subsubsection{ChatGPT4}
As a proxy for neural spell-checker, we employ ChatGPT powered by the GPT-4 model, specifically the \textquotedblleft gpt-4-0125-preview\textquotedblright\ version.

In structuring prompts for ChatGPT, we first clearly define the problem by directing ChatGPT to detect and highlight specific spelling mistakes in a given text. We enhance the prompt with one generic example that includes misspelled words and words incorrectly written together or apart. 
We instruct ChatGPT to tag these errors using three distinct labels (\textquotedblleft \textless mistake\textgreater\textquotedblright, \textquotedblleft \textless split\textgreater\textquotedblright\  and \textquotedblleft \textless concat\textgreater\textquotedblright). Later, during model output post-processing, we merge these labels into a single category. Our preliminary experimentation indicates that requesting labels for each type of mistake leads to enhanced performance. 

\subsection{Analysis of results}
We first analyze the quantitative results (shown in Table \ref{tab:f05scores}) using the $F_{0.5}$ score as the metric, as is common in grammatical error correction literature. The difference to the better known $F_1$ score is that $F_{0.5}$ score gives more emphasis to precision than to recall; this is a natural choice for spelling detection systems, where the importance of correct error identification is greater than capturing all errors.

\begin{table}[htb]
\centering
\caption{The evaluation results for different systems on the three Slovene datasets. We report macro $F_{0.5}$ scores. }
\label{tab:f05scores}
\begin{tabular}{lccc}
\toprule
\textbf{System} & \textbf{Synthetic-Eval} & \textbf{Šolar-Eval} & \textbf{Lektor-Spelling} \\
\midrule
HunSpell\textsubscript{SL} & $0.63$ & $0.53$ & $0.51$ \\
LanguageTool\textsubscript{SL} & $0.84$ & $0.75$ & $0.63$ \\
ChatGPT4\textsubscript{1S} & $0.83$ & $0.65$ & $0.57$ \\
SloSpell & $0.88$ & $0.79$ & $0.62$ \\
SloNSpell & $\textbf{0.97}$ & $\textbf{0.92}$ & $\textbf{0.65}$ \\
\bottomrule
\end{tabular}
\end{table}

The results show the superiority of the proposed neural spell-checker SloNSpell over the tested baselines on all three benchmarks. 
Surprisingly, our lexicon-based SloSpell spell-checker achieves second or third best scores, showcasing the strength of a simple spell-checker using a large, high-quality, curated lexicon. 
Most notably, SloSpell and LanguageTool both surpass ChatGPT4 in the one-shot learning setting. The poor performance of ChatGPT4 on Šolar-Eval can potentially be attributed to the relatively specific vocabulary used in school essays (e.g., names of characters in Slovene literary works) to which ChatGPT4 was not exposed during its training. The lower score on the Synthetic-Eval dataset may be caused by the synthetic errors resulting in very unnatural words, whereas ChatGPT4 has presumably only observed a tiny portion of (more) natural Slovene text.
The worst scores overall are achieved by the HunSpell\textsubscript{SL} spell-checker, likely due to the relatively small and outdated Slovene dictionary.
Time-wise, SloNSpell is computationally much more demanding than SloSpell -- on Šolar-Eval, SloNSpell processes approximately three examples per second (running on an Apple M3 Pro GPU) while SloSpell processes approximately $250$ examples per second (running on an Intel i7-1165G7 CPU).
Next, we qualitatively analyze the results to outline the strengths and weaknesses of the approaches.

We analyzed the predictions made by the top-performing system, SloNSpell, with a primary focus on its false positive and false negative predictions, and to a lesser extent, its true positive predictions, through a qualitative assessment across the Šolar-Eval (Š-E) and Lektor-Spelling (L-S) datasets. In Š-E, we examined 907 predictions, while in L-S, we reviewed 355 predictions. Notably, the categories below reflect the distinctions between the text types in both datasets (e.g., error types typical for learner texts in Š-E versus the presence of rare, specialized vocabulary in L-S's technical texts) and differences stemming from error annotation methods. Some identified issues with evaluation datasets will be addressed in our future development.

The occurrence of false negatives (FNs) in both datasets is comparable: 152 (16.8\%) in Š-E and 64 (18\%) in L-S. Within this category, misspelled \textit{proper names} are highly typical for learner texts (32.2\%) but are absent in the L-S dataset. Both datasets feature falsely accepted misspelled forms that are formally \textit{identical to another legitimate word form} (23.7\% in Š-E and 14.1\% in L-S). Spell-checkers also struggle with \textit{language variants} that are acceptable in certain contexts but not in others (5.3\% FNs in Š-E and 1.6\% in Lektor-Spelling). Unique to the L-S annotation are FNs related to ambiguous instances of \textit{writing words separately or together} (20.3\%) and the occurrence of \textit{redundant, repeated words} (18.8\%). These FN types are traditionally challenging for spell-checking and are anticipated by potential users of the system. However, some unexpected types were also discovered, ranging from falsely accepted \textit{foreign words} (4.7\% in L-S) and \textit{archaic forms} (2\% in Š-E and 3.1\% in L-S) to completely \textit{non-existent forms} (16.4\% in Š-E and 6.3\% in L-S). Similarly perplexing is the treatment of the Slovene \textit{preposition s/z}, which is selected according to the phonetic form of the following word. Despite the relative simplicity of the selection rules, the system rather randomly errs in this regard (9.9\% in Š-E).

The evaluation of false positives (FPs) also revealed both expected and unexpected categories. In Šolar-Eval, 118 (13\%) and in L-S, 229 (64.5\%) FPs were found. \textit{Proper names} were often misidentified as misspelled (53.4\% in Š-E and 21.4\% in L-S). \textit{Rare words or forms }presented another expected category, with 7.6\% in Š-E and 23.1\% in L-S (mainly technical terminology). Similarly to FNs, problems pertaining to \textit{language variants} (5.1\% in Š-E and 1.7\% in L-S), words of \textit{foreign origin} (1.7\% in L-S), and \textit{ambiguous word separation} (15.3\% in Š-E and 2.6\% in L-S). Unique to the L-S annotation are FNs related to \textit{numerals, special symbols} (9.6\%), \textit{single letters} (5.7\%), \textit{abbreviations} (8.7\%), and \textit{word capitalization} (4.8\%). Again, there are unexpected issues with the \textit{preposition s/z} (0.9\% in L-S) and a considerately large group of frequently occurring words that were flagged as problematic for \textit{inexplicable reasons }(16.9\% in Š-E and 10.5\% in L-S).

Contrary to false positives, true positives were more frequent in Šolar-Eval, with 637 instances (70.2\%), compared to Lektor-Spelling, which had 62 instances (17.5\%). Most of the categories mentioned with FNs and FPs can also be found here, with correct predictions by the system (96.4\% in Š-E and 88.7\% in L-S). 

A set of issues arose from the internal composition of Lektor-Spelling, notably including examples of transcribed spoken language, which represented 4.8\% of TPs, 23.4\% of FNs, and 9.2\% of FPs. These instances differ significantly from the dataset's intended genre and should be removed in future revisions. A smaller proportion of TPs (3.6\% in Š-E and 6.5\% in L-S), FNs (2.6\% in Š-E and 4.7\% in L-S), and FPs (1.7\% in Š-E) can be attributed to dataset issues such as extraction and annotation errors. Moreover, a significant portion of FNs (7.9\% in Š-E and 3.1\% in L-S) revealed that non-standard word forms were included in the Sloleks lexicon without proper labeling. We will address all these issues in our forthcoming work.

Finally, the analysis revealed that specific words, such as proper names, might be accepted in one instance and flagged in another within the same context. From the user's perspective, this undermines the system's reliability and is thus a main priority for further improvement.

\section{Conclusion}
\label{sec:conclusion}

In this study, we created two new spell-checkers for Slovene: SloSpell, a classical approach with a considerably larger vocabulary compared to existing ones, and SloNSpell, a novel neural approach. To train SloNSpell, we developed a novel synthetic data creation tool. We also tested two new evaluation datasets, a synthetically generated one and the other from the Lektor corpus, which we have fully annotated with spelling errors. We compared the newly created methods with three existing classical and one neural approach. 

The SloNSpell model, a fine-tuned BERT model, outperformed all other methods, achieving the highest $F_{0.5}$ scores across the three evaluation datasets. This model excelled on the Synthetic-Eval and Šolar-Eval datasets, with $F_{0.5}$ scores exceeding 0.9, but with lower performance on the Lektor dataset, due to the presence of error types not covered in the training data, such as word duplication, unnecessary words, and slang.

In further work, the identified weaknesses will be used to form additional types of training data. Refining ChatGPT's prompts could boost its now disappointing performance. However, considering the cost and speed, the SloNSpell approach is likely to remain superior for the spelling correction task, while SloSpell's large vocabulary shall replace existing Slovene vocabulary in various tools.

\section*{Acknowledgments}
The work was partially supported by the Slovenian Research and Innovation Agency (ARIS) core research programme P6-0411, a young researcher grant, as well as projects J7-3159, CRP V5-2297, and L2-50070. 

\bibliographystyle{unsrt}  
\bibliography{templateArxiv}

\begin{thebibliography}{10}

\bibitem{sloleks3}
Jaka {\v C}ibej, Kaja Gantar, Kaja Dobrovoljc, Simon Krek, Peter Holozan, Toma{\v z} Erjavec, Miro Romih, {\v S}pela Arhar~Holdt, Luka Krsnik, and Marko Robnik-{\v S}ikonja.
\newblock Morphological lexicon {Sloleks} 3.0, 2022.
\newblock {S}lovenian language resource repository {CLARIN}.{SI}.

\bibitem{solar-eval}
{\v S}pela Arhar~Holdt, Polona Gantar, Mija Bon, Magdalena Gapsa, Polona Lavri{\v c}, and Matej Klemen.
\newblock Dataset for evaluation of {Slovene} spell- and grammar-checking tools {Šolar-Eval 1.0}, 2023.
\newblock {S}lovenian language resource repository {CLARIN}.{SI}.

\bibitem{dcspell-chinese-electra}
Jing Li, Gaosheng Wu, Dafei Yin, Haozhao Wang, and Yonggang Wang.
\newblock {DCSpell}: A detector-corrector framework for {C}hinese spelling error correction.
\newblock In {\em Proceedings of the 44th International ACM SIGIR Conference on Research and Development in Information Retrieval}, page 1870–1874, 2021.

\bibitem{clark2020electra}
Kevin Clark, Minh-Thang Luong, Quoc~V. Le, and Christopher~D. Manning.
\newblock {ELECTRA}: Pre-training text encoders as discriminators rather than generators.
\newblock In {\em ICLR}, 2020.

\bibitem{ji2021-spellbert}
Tuo Ji, Hang Yan, and Xipeng Qiu.
\newblock {S}pell{BERT}: A lightweight pretrained model for {C}hinese spelling check.
\newblock In Marie-Francine Moens, Xuanjing Huang, Lucia Specia, and Scott Wen-tau Yih, editors, {\em Proceedings of EMNLP 2021}, pages 3544--3551, November 2021.

\bibitem{croatian-err-corr}
Maja Mitreska, Kostadin Mishev, and Monika Simjanoska.
\newblock {NLP}-based typo correction model for {C}roatian language.
\newblock In {\em 2022 45th Jubilee International Convention on Information, Communication and Electronic Technology (MIPRO)}, pages 942--947, 2022.

\bibitem{urdu-spelling-err-det-corr}
Romila Aziz, Muhammad~Waqas Anwar, Muhammad~Hasan Jamal, and Usama~Ijaz Bajwa.
\newblock A hybrid model for spelling error detection and correction for {U}rdu language.
\newblock {\em Neural Computing and Applications}, 33(21):14707--14721, Nov 2021.

\bibitem{erjavec2003-mte}
Toma{\v{z}} Erjavec, Cvetana Krstev, Vladim{\'\i}r Petkevi{\v{c}}, Kiril Simov, Marko Tadi{\'c}, and Du{\v{s}}ko Vitas.
\newblock The {MULTEXT}-east morphosyntactic specification for {S}lavic languages.
\newblock In {\em Proceedings of the 2003 {EACL} Workshop on Morphological Processing of {S}lavic Languages}, pages 25--32, April 2003.

\bibitem{sloberta}
Matej Ul{\v{c}}ar and Marko Robnik-{\v{S}}ikonja.
\newblock {SloBERTa: Slovene} monolingual large pretrained masked language model.
\newblock {\em Proceedings of SI-KDD within the Information Society 2021}, pages 17--20, 2021.

\bibitem{Vaswani2017}
Ashish Vaswani, Noam Shazeer, Niki Parmar, Jakob Uszkoreit, Llion Jones, Aidan~N Gomez, {\L}ukasz Kaiser, and Illia Polosukhin.
\newblock Attention is all you need.
\newblock In {\em Advances in neural information processing systems}, pages 5998--6008, 2017.

\bibitem{krek-etal-2020-gigafida}
Simon Krek, {\v{S}}pela Arhar~Holdt, Toma{\v{z}} Erjavec, Jaka {\v{C}}ibej, Andraz Repar, Polona Gantar, Nikola Ljube{\v{s}}i{\'c}, Iztok Kosem, and Kaja Dobrovoljc.
\newblock Gigafida 2.0: The reference corpus of written standard {S}lovene.
\newblock In {\em Proceedings of LREC 2020}, pages 3340--3345, May 2020.

\bibitem{solar3}
{\v S}pela Arhar~Holdt, Tadeja Rozman, Mojca Stritar~Ku{\v c}uk, Simon Krek, Irena Krap{\v s}~Vodopivec, Marko Stabej, Eva Pori, Teja Goli, Polona Lavri{\v c}, Cyprian Laskowski, Polonca Kocjan{\v c}i{\v c}, Bojan Klemenc, Luka Krsnik, and Iztok Kosem.
\newblock Developmental corpus {\v s}olar 3.0, 2022.
\newblock {S}lovenian language resource repository {CLARIN}.{SI}.

\bibitem{gantar2023solar-eval}
Polona Gantar, Mija Bon, Magdalena Gapsa, and {\v{S}}pela Arhar~Holdt.
\newblock {Šolar-Eval: Evalvacijska} množica za strojno popravljanje jezikovnih napak v slovenskih besedilih.
\newblock {\em Jezik in slovstvo}, 68(4):89--108, 2023.

\bibitem{lektor}
Damjan Popi{\v{c}}.
\newblock Revising translation revision in {Slovenia}.
\newblock {\em New Horizons in Translation Research and Education 2}, page~72, 2014.

\end{thebibliography}

\end{document}